\documentclass[a4paper]{article}

\usepackage{INTERSPEECH2020}

\usepackage{calrsfs}
\usepackage{calc}
\usepackage{amsmath,amssymb,pifont}
\usepackage{multirow}
\usepackage{enumitem}

\usepackage{color,soul}

\title{Augmenting Images for ASR and TTS through\\Single-loop and Dual-loop Multimodal Chain Framework}
\name{Johanes Effendi$^{1,2}$, Andros Tjandra$^1$, Sakriani Sakti$^{1,2}$, Satoshi Nakamura$^{1,2}$}
\address{
	$^1$Nara Institute of Science and Technology, Japan\\
	$^2$RIKEN, Center for Advanced Intelligence Project (AIP), Japan}
\email{\{johanes.effendi.ix4, andros.tjandra.ai6, ssakti, s-nakamura\}@is.naist.jp}

\begin{document}
	
	\maketitle
    \vspace{-0.15cm}
	\begin{abstract}
    \vspace{-0.2cm}
		Previous research has proposed a machine speech chain to enable automatic speech recognition (ASR) and text-to-speech synthesis (TTS) to assist each other in semi-supervised learning and to avoid the need for a large amount of paired speech and text data. However, that framework still requires a large amount of unpaired (speech or text) data. A prototype multimodal machine chain was then explored to further reduce the need for a large amount of unpaired data, which could improve ASR or TTS even when no more speech or text data were available.
		Unfortunately, this framework relied on the image retrieval (IR) model, and thus it was limited to handling only those images that were already known during training. Furthermore, the performance of this framework was only investigated with single-speaker artificial speech data. In this study, we revamp the multimodal machine chain framework with image generation (IG) and investigate the possibility of augmenting image data for ASR and TTS using single-loop and dual-loop architectures on multispeaker natural speech data. Experimental results revealed that both single-loop and dual-loop multimodal chain frameworks enabled ASR and TTS to improve their performance using an image-only dataset.
	\end{abstract}
	
	\vspace{-0.2cm}
	\section{Introduction}
	\vspace{-0.2cm}
	Machines need to learn how to listen or speak in order to communicate with humans. Traditionally, learning to listen and speak is done through the development of automatic speech recognition (ASR) and text-to-speech synthesis (TTS) that are trained separately and independently using a supervised method. Such training requires a large amount of paired data consisting of speech and corresponding transcriptions, which is often unavailable. When the training is finished, the machine is only able to speak or listen to others, but it still cannot hear its own voice.
	
	Humans, by contrast, have a closed-loop speech chain mechanism with auditory feedback from a speaker's mouth to her ears \cite{denes1993speech}. By simultaneously listening and speaking, the speaker can monitor her volume and articulation and can improve the general comprehensibility of her speech. Furthermore, overall human communication channels include not only auditory channels but also visual channels. Having multiple information sources that are perceived together builds general concepts and understanding.
	
\begin{figure*}[!htbp]
		\centering
		\includegraphics[width=1.9\columnwidth]{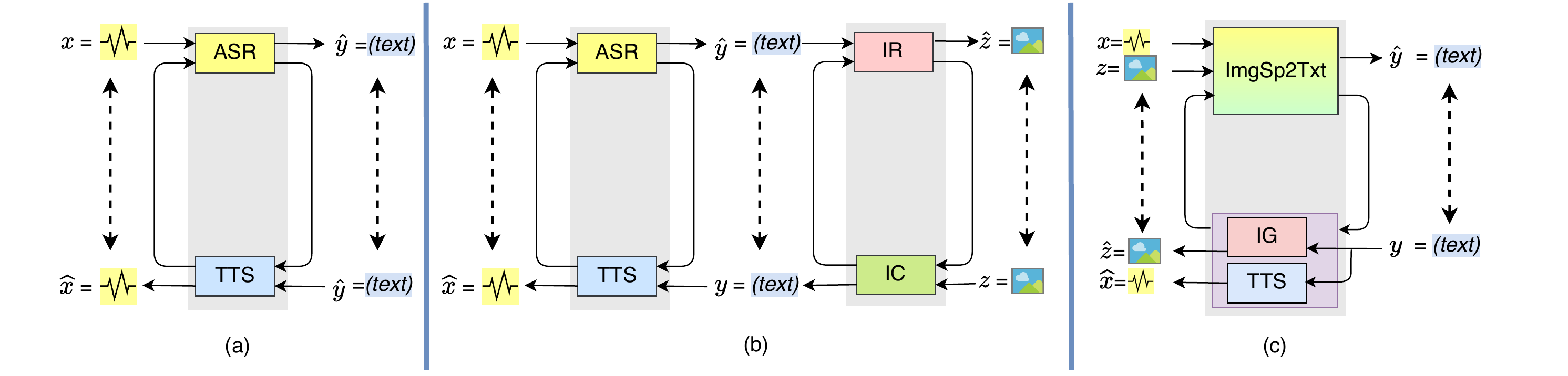}
		\vspace{-0.4cm}
		\caption{Architecture: (a) machine speech chain \cite{tjandra_schain1}, (b) previous multimodal machine chain \cite{effendi2019speech} (MMC1) and (c) proposed multimodal machine chain (MMC2).}
		\label{fig:chain}
		\vspace{-0.4cm}
	\end{figure*}

	Inspired by this human mechanism, a machine speech chain based on deep learning \cite{tjandra_schain4,tjandra_schain1,tjandra_schain2,tjandra_schain3} was previously proposed to create a machine that can learn not only to listen or speak but also to listen while speaking. The advantages of the framework shown in Fig.~\ref{fig:chain}(a) are that it enables ASR and TTS to assist each other in semi-supervised learning and it avoids the need for a large amount of paired speech and text data. However, the framework still requires a large amount of unpaired (speech or text) data.
	
	A multimodal machine chain \cite{effendi2019speech} was then proposed to mimic overall human communication and accommodate a visual modality on top of speech and text modalities. This framework (see Fig.~\ref{fig:chain}(b)) with a dual-loop architecture consists of: (1) the original speech chain with ASR and TTS; and (2) an additional visual chain component with image captioning (IC) and image retrieval (IR). The results revealed that this framework further reduced the need for unpaired data, as it could improve ASR or TTS even when no more  speech and text data were available. However, as the framework relied on the IR model, it could only handle those images that were already known during training. Another limitation was that the performance was only investigated on single-speaker synthesized speech data that were generated using Google TTS.
	
	In this study, we revamp the multimodal machine chain with image generation (IG). We also explore an alternative single-loop mechanism with a multisource architecture (see Fig.~\ref{fig:chain}(c)), based on the assumption that humans do not separate audio and visual information when perceiving it. Furthermore, we investigate the performance of the proposed framework on multispeaker natural speech data and show the possibility of augmenting image data for further enhancing ASR and TTS. Here, the upgraded dual-loop multimodal chain is denoted as MMC1, and the new single-loop multimodal chain is denoted as MMC2.
	
	\vspace{-0.2cm}
	\section{Multimodal Machine Chain}
	\vspace{-0.25cm}
	\subsection{Overview}
	\label{ssec:overview}
	\vspace{-0.2cm}
	
	The architecture of MMC1 is similar to the one illustrated in Fig.~\ref{fig:chain}(b) but replacing the IR with IG. The new MMC2 framework, shown in Fig.~\ref{fig:chain}(c), merges the speech chain and visual chain into a single-loop mechanism. Consequently, ASR and IC are now combined into ImgSp2Txt with a dual-decoder so that the mechanism can disambiguate input using the fusion of image and speech. MMC2 provides a more compact architecture than MMC1, and with only one loop chain, the framework is similar to the original architecture of the machine speech chain. Our motivation in proposing MMC2 is to show that the multimodal chain training strategy can also be applied for a multi-source multimodal model such as audio-visual ASR (ImgSp2Txt).
	
	There are various ways to train MMC1 and MMC2 depending on the availability of the data, including speech (x), text (y), and images (z) in paired (P) or unpaired (U) conditions. When all paired data (speech-image-text) are available, all components (ImgSp2Txt, ASR, IC, TTS, and IG) can be trained independently in supervised training by minimizing the loss between their predicted target sequence and the ground truth sequence. However, when only unpaired data are available, ImgSp2Txt, ASR, IC, TTS, and IG need to assist each other in unsupervised training through a loop connection. To further clarify the learning process during unsupervised training in the MMC1 and MMC2 frameworks, we unrolled the architecture to describe the training mechanism:
	
	\begin{enumerate}
		\item \textbf{Speech and/or Image to Text} \\
		\vspace{-0.65cm}
		\begin{figure}[!htbp]
			\centering
			\includegraphics[width=0.80\columnwidth]{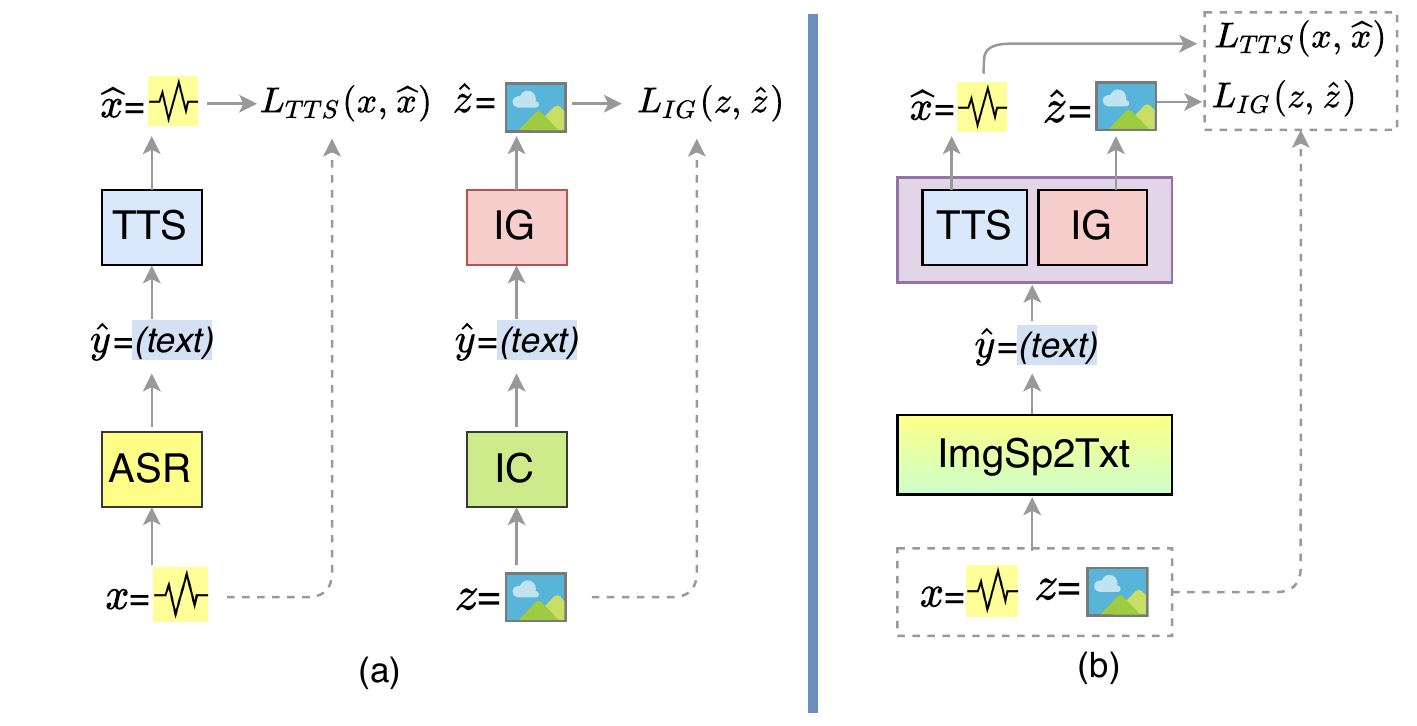}
			\vspace{-0.5cm}
			\caption{Unrolled process from speech and/or image to text in (a) MMC1 and (b) MMC2.}
			\label{fig:av_to_txt}
		\end{figure}
		\vspace{-0.3cm}
		
		Given only speech $x$ and/or image $z$ data (without any corresponding transcription $y$), in MMC1 two separate unrolled processes are needed for the speech chain and the visual chain (see Fig.~\ref{fig:av_to_txt}(a) left and right, respectively). In contrast, with MMC2, the unrolled process from speech and/or image to text can be done simultaneously (see Fig.~\ref{fig:av_to_txt}(b)). ImgSp2Txt transcribes the input of either $\{xz, x, z\}$ into text $\hat{y}$. This generated text is then used by TTS to synthesize speech $\hat{x}$ and/or by IG to generate an image $\hat{z}$. By comparing both the original and the predicted results, TTS and IG parameters can be updated with backpropagation by minimizing the loss $L_{TTS}(x, \hat{x})$ and $L_{IG}(z, \hat{z})$, respectively.
		\vspace{-0.1cm}
		\begin{equation}		
		\hat{x} = TTS(\text{\textit{ImgSp2Txt}}(x,z))
		\vspace{-0.1cm}
		\end{equation}
		\begin{equation}		
		\hat{z} = IG(\text{\textit{ImgSp2Txt}}(x,z))
		\vspace{-0.1cm}	
		\end{equation}
		\begin{equation}
		\theta_{TTS} = Optim(\theta_{TTS}, \bigtriangledown_{\theta_{TTS}L_{TTS}(x,\hat{x})}) 	
		\vspace{-0.1cm}
		\end{equation}
		\begin{equation}
		\theta_{IG} = Optim(\theta_{IG}, \bigtriangledown_{\theta_{IG}L_{IG}(z,\hat{z})}) 	
		\vspace{-0.1cm}
		\end{equation}
		
		\vspace{0.1cm}
		\item \textbf{Text to Speech and/or Image} \\
		\vspace{-0.3cm}
		\begin{figure}[h]
			\vspace{-0.2cm}
			\centering
			\includegraphics[width=0.80\columnwidth]{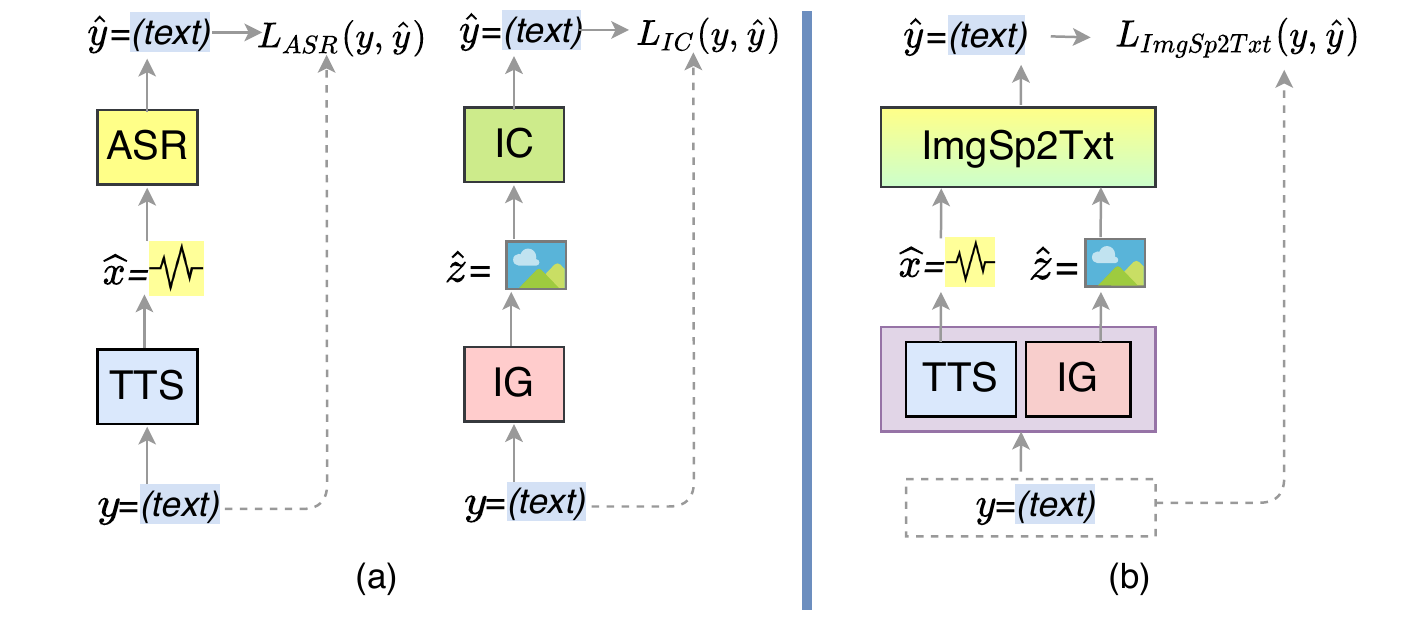}
			\vspace{-0.4cm}
			\caption{Unrolled process from text to speech and/or image in (a) MMC1 and (b) MMC2.}
			\label{fig:txt_to_av}	
			\vspace{-0.2cm}
		\end{figure}
		
		Here, only text $y$ is available, but none of the speech $x$ and image $z$ data are available. In such a condition, speech $\hat{x}$ can be generated using TTS and image $\hat{z}$ can be generated using IG. In the MMC1 framework as illustrated in Fig.~\ref{fig:txt_to_av}(a), the predicted speech $\hat{x}$ and image $\hat{z}$ are used separately by the ASR in the speech chain and the IC in the visual chain, respectively. By contrast, in the MMC2 framework shown in Fig.~\ref{fig:txt_to_av}(b), ImgSp2Txt can use both predicted speech $\hat{x}$ and image $\hat{z}$ together and decode a text hypothesis $\hat{y}$. By measuring the reconstruction loss between $y$ and $\hat{y}$, ImgSp2Txt parameters are updated with backpropagation, and the loss $L_{ImgSp2Txt}(y,\hat{y})$ is minimized.
		\vspace{-0.1cm}
		\begin{equation}		
		\hat{y} = \text{\textit{ImgSp2Txt}}(TTS(y),IG(y))
		\vspace{-0.1cm}		
		\end{equation}
		\begin{equation}
		\begin{split}
		\theta_{\text{\textit{ImgSp2Txt}}} = Optim(\theta_{\text{\textit{ImgSp2Txt}}}, 	
		\bigtriangledown_{\theta_{\text{\textit{ImgSp2Txt}}}L_{\text{\textit{ImgSp2Txt}}}(y,\hat{y})}) 	
		\vspace{-0.1cm}
		\end{split}
		\end{equation}
	\end{enumerate}
	
	\vspace{-0.25cm}
	\subsection{Multimodal Chain Components}
	\label{ssec:comp}
	\vspace{-0.55cm}
	
	\begin{figure}[!h]
		\centering
		\includegraphics[width=0.75\columnwidth]{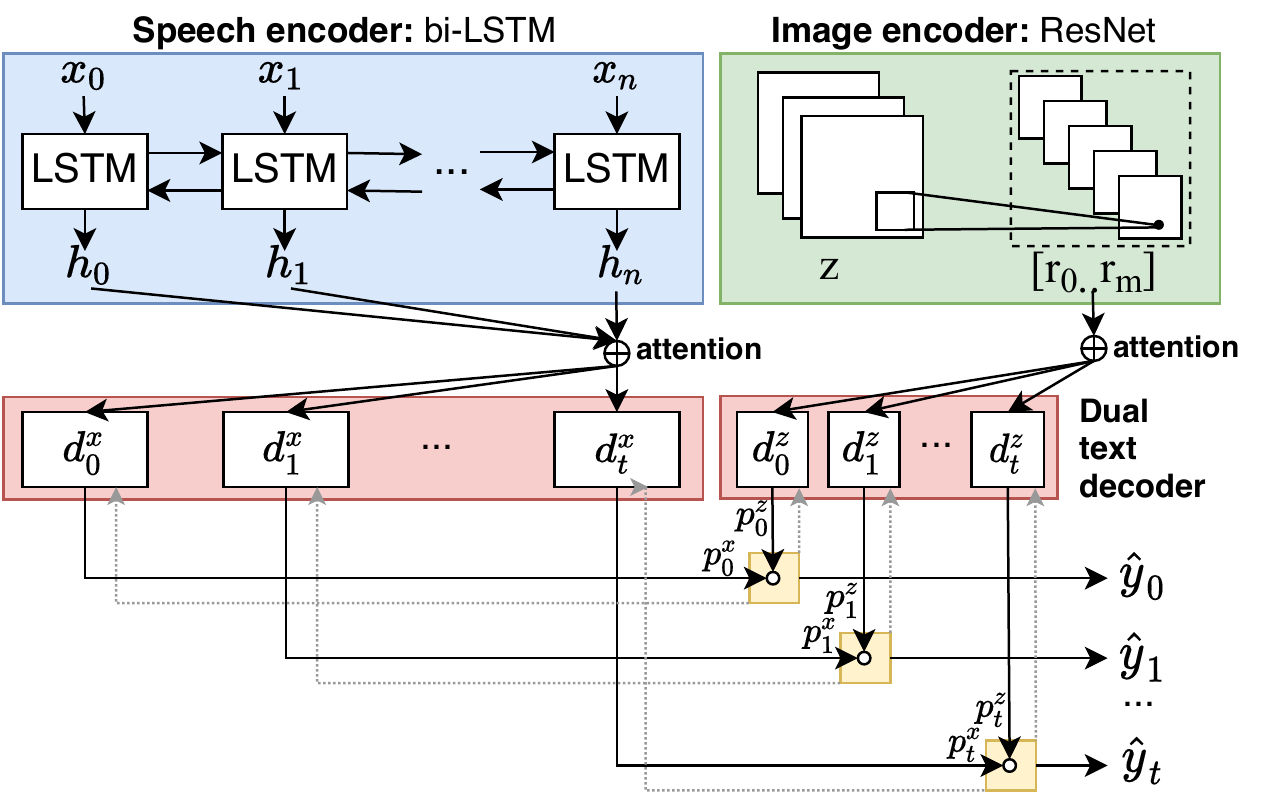}
		\vspace{-0.3cm}
		\caption{Dual text-decoder that combines audio and visual information.}
		\label{fig:avasr}	
	\end{figure}
    \vspace{-0.3cm}
	
	The MMC1 framework has four components: (1) ASR-based on `Listen, Attend, and Spell (LAS)' architecture
	\cite{chan2016listen}; (2) TTS-based Tacotron \cite{wang_taco} with slight modification as previously reported \cite{tjandra_schain1}; (3) Attentional-based IC model \cite{xu2015show}; and (4) IG, in which, in this study, attentional generative adversarial networks (AttnGAN) \cite{attngan} are used.
	
	By contrast, in the MMC2 framework, instead of having ASR and IC, a sequence-to-sequence ImgSp2Txt with dual-decoder is used to perform audiovisual to text tasks (see Fig.~\ref{fig:avasr}). Here, the speech encoder consists of a bidirectional LSTM that encodes Mel-spectrogram $[x_0..x_n]$ into encoded representation $[h_0..h_n]$, and the image encoder produces high-level feature representation $[r_0..r_m]$ based on a residual network (ResNet) \cite{he2016deep}. Then, the dual text decoder $d_t^x$ and $d_t^z$ attend the $[h_0..h_n]$ and $[r_0..r_m]$ respectively. During training, when both image and speech are available, the calculation of the $L_{ImgSp2Txt}$ loss are calculated by taking average of both $p_t^x$ and $p_t^z$ output layer probability. However, when only images or speech are available, then the decoder uses only the corresponding output layer. Note that, compared with IC, the image captioning in ImgSp2Txt works at the character level to follow the best practices in ASR.
	
	The TTS and IG components on MMC2 are the same as in MMC1. In addition, as the performance will be investigated on multispeaker natural speech data, we constructed an additional element for speaker recognition (SPKREC), which is based on DeepSpeaker \cite{deepspeaker}. To handle unseen speakers, the TTS will perform one-shot speaker adaptation as previously reported \cite{tjandra_schain2}.
	
	In practice, to reduce memory usage, both IC and IG used only a 128x128 image size. SPKREC followed the original hyper-parameters in the previous paper \cite{tjandra_schain2}, except the embedding size was reduced from 128 to 64. We decoded all hypotheses during semi-supervised training and testing using a beam size of three. We used an Adam optimizer with a learning rate of 1e-4 for the ASR, ImgSp2Txt, and IC models, 2.5e-4 for the TTS model, and 2e-4 for the IG model.
	
	\vspace{-0.2cm}
	\section{Experimental Set-up}
	\label{sec:setup}
	\vspace{-0.1cm}
	
	We ran our experiment with the Flickr8k dataset \cite{flickr8k_corpus}, which has 8000 photos of everyday activities, events, and scenes. Each image has five captions that describe the image with a vocabulary of 8920 words. Harwath and Glass (2015) \cite{flickr8k_audio} extended this corpus using Amazon Mechanical Turk to collect 40000 corresponding examples of natural speech by 183 different speakers for a total of 64 speech hours.
	
	\begin{table}[b]
		\vspace{-0.3cm}	
		\footnotesize
		\centering
		\caption{Training data partition for Flickr8k with three conditions: (1) available paired data, denoted as $\bigcirc$, (2) available but unpaired data, denoted as $\blacktriangle$, and unavailable data, denoted as $\times$. Each image has five captions and five speech utterances.}	
		\vspace{-0.3cm}
		\label{tab:data}
		\begin{tabular}{c|cccr}
			\multirow{2}{*}{\textbf{Dataset}} & \textbf{Speech}   & \textbf{Text}    & \textbf{Image}   & \multirow{2}{*}{\textbf{\# Image}} \\
			& \textbf{$x$}      & \textbf{$y$}     & \textbf{$z$}     &  \\ \hline
			Multimodal (Paired)    & $\bigcirc$        & $\bigcirc$       & $\bigcirc$       & 800                \\
			Multimodal (Unpaired)    & $\blacktriangle$  & $\blacktriangle$ & $\blacktriangle$ & 1500              \\
			Speech only (Unpaired)        & $\blacktriangle$  & $\times$         & $\times$         & 1850            \\
			Image only (Unpaired)        & $\times$          & $\times$         & $\blacktriangle$ & 1850             \\
		\end{tabular}
		\vspace{-0.2cm}
	\end{table}
	
	\begin{table}[b]
		\vspace{-0.3cm}
		\footnotesize
		\centering
		\caption{Our IG and ImgSp2Txt performances compared with existing published results.}
		\vspace{-0.3cm}
		\label{tab:topline}
		\begin{tabular}{|ll|c|}
			\hline
			\textbf{Data}&{\textbf{Model}}&{\textbf{Result}}\\
			\hline
			\hline
			\multicolumn{3}{|c|}{\textbf{IG - IS (\%) $\uparrow$}}\\
			CUB & Xu et al. \cite{attngan} & 4.36 \\
			& Ours (MMC1) & 5.67 \\
			& Ours (MMC2) & 5.67 \\
			\hline
			\multicolumn{3}{|c|}{\textbf{ImgSp2Txt - CER / WER (\%) $\downarrow$}}\\
			Flickr8k & Sun et al. \cite{sun_f8k} & - / 13.81 \\
			& Ours (MMC1) & 5.76 / 9.76 \\
			& Ours (MMC2) & 5.16 / 7.13 \\
			\hline
		\end{tabular}
		\vspace{-0.2cm}	
	\end{table}

	\begin{table*}[!htb]
		\vspace{-0.3cm}
		\footnotesize
		\centering
		\caption{Performance of proposed MMC1 and MMC2 compared with label propagation method in Flickr8k dataset based on training set-up in Table~\ref{tab:data} (P=paired data; U=unpaired data). The last line is the topline system when the 6k images with the corresponding five captions and five speech utterances for each image are available.}
		\vspace{-0.35cm}
		\label{tab:result_audiovisual}
		\begin{tabular}{clr|cccc|cccc}
			\multirow{3}{*}{\textbf{Training}} & \multirow{3}{*}{\textbf{Data Type}} & \multirow{3}{*}{\textbf{\#Image}} & \multicolumn{4}{c|}{MMC1} & \multicolumn{4}{c}{MMC2}  	\\
			& & & \textbf{ASR} & \textbf{IC} & \textbf{TTS} & \textbf{IG} & \multicolumn{2}{c}{\textbf{ImgSp2Txt}} & \textbf{TTS} & \textbf{IG}\\
			& & & CER$\downarrow$ & B4$\uparrow$ & L2$^2$$\downarrow$ & IS$\uparrow$ &CER$\downarrow$ & B4$\uparrow$ & L2$^2$$\downarrow$ & IS$\uparrow$\\ \hline
			\textbf{Label Propagation I} & Multimodal (P) & 800 & 36.35 & 12.75 & 0.77 & 5.90 & 26.67 & 32.23 & 0.77 & 5.90 \\
			\textbf{(Semi-Supervised)} & + Multimodal (U) & 1500 & 39.57 & 12.53 & 0.77 & 7.04 & 27.45 & 33.59 & 0.77 & 7.04\\
			\textbf{} & + Sp only (U) & 1850 & 46.04 & - & 0.63 & - & 28.87 & 35.75 & 0.63 & - \\
            \textbf{} & + Img only (U) & 1850 & - & 11.41 & - & 7.20 & 30.31 & 35.38 & - & 7.20 \\
			\hline
			\textbf{Label Propagation II} & Multimodal (P) & 800+$\alpha$ & 15.52 & 15.10 & 0.64 & 7.25 & 13.54 & 57.63 & 0.64 & 7.25 \\
			\textbf{Plus $\alpha=600$} & + Multimodal (U) & 1500-$\alpha$ & 15.36 & 15.63 & 0.62 & 7.82 & 13.22 & 58.66 & 0.62 & 7.82\\
			\textbf{(Semi-Supervised)} & + Sp only (U) & 1850 & 15.28 & - & 0.55 & - & 14.36 & 59.36 & 0.55 & - \\
            \textbf{} & + Img only (U) & 1850 & - & 15.86 & - & 8.86 & 15.24 & 58.69 & - & 8.86 \\
			\hline \hline
			\textbf{Proposed} & Multimodal (P) & 800 & 36.35 & 12.75 & 0.77 & 5.90 & 26.67 & 32.23 & 0.77 & 5.90 \\
			\textbf{Multimodal Chain} & + Multimodal (U) & 1500 & 15.10 & 13.22 & 0.59 & 8.29 & 14.88 & 55.15 & 0.65 & 10.12\\
			\textbf{(Semi-Supervised)} & + Sp only (U) & 1850 & 12.37 & 13.28 & 0.56 & 9.12 & 13.81 & 58.03 & 0.62 & 10.65\\
			 & + Img only (U) & 1850 & 12.06 & 13.29 & 0.56 & 9.11 & 12.32 & 59.66 & 0.61 & 9.95\\
			\hline
			\textbf{Topline (Supervised)} & Multimodal (P) & 6000 & 5.76 & 19.91 & 0.50 & 9.66 & 5.16 & 79.88 & 0.50 & 9.66\\
			\hline
		\end{tabular}
		\vspace{-0.35cm}
	\end{table*}
	
	The Flickr8k data is commonly divided into 6k training images, 1k validation images, and 1k test images. In an ideal situation, 6k images with corresponding five captions and five speech utterances for each image can be used for training. However, in reality, a large amount of paired data is often unavailable. Therefore, in this study, we investigated the capability of our proposed framework given a non-ideal situation. We formulated training sets, as listed in Table~\ref{tab:data}. Among the 6k data, only 800 units had complete speech+image+text paired data. In another 1500 units, there were speech, image, and text data, but they were unpaired or not related to each other. Then, in the remaining data, 1850 units had only speech utterances, and 1850 units had only image data.
	
	Despite having such limited data, our proposed framework could use all of those data in a semi-supervised setting. First, we trained all model components independently in supervised training with 800 (speech+image+text) paired units. Then, we further enhanced the performance and continued the training process using the 1500 multimodal (speech/image/text) unpaired data. Here, those model components needed to support each other in unsupervised training through a loop connection, as described in Section~\ref{ssec:overview}. Specifically, we performed two unrolled processes separately: (1) speech/image to text, and (2) text to speech/image. In MMC1, this training step is equivalent to the training speech chain (ASR and TTS) and the visual chain (IC and IG) separately (see Fig.~\ref{fig:av_to_txt}(a) and Fig.~\ref{fig:txt_to_av}(a)). But, in MMC2, this training step is a semi-supervised single loop chain between ImgSp2Txt, TTS, and IG (see Fig.~\ref{fig:av_to_txt}(b) and Fig.~\ref{fig:txt_to_av}(b)). Finally, given the remaining single modality data (1850 units with speech only or 1850 units with image only), we further trained the components by using both unrolled processes together. We performed the speech/image to text unrolled process, and we then used the predicted text to perform the text to speech/image unrolled process. This way, we were still able to enhance ASR and TTS given only image data. Such a mechanism cannot be done with the original machine speech chain. As a comparison, we also performed another semi-supervised technique, the label propagation method \cite{labelprop}. Here, we used the initial models to generate the missing information and retrained the model using supervised learning.
	
	For evaluation metrics, we measured the performance of the ASR using a character error rate or word error rate (CER/WER), and a bilingual evaluation understudy (BLEU) \cite{papineni2002bleu} for the IC, which compares matching n-grams between a hypothesis and a reference sentence (higher being better). Specifically, we used BLEU4 with 4-grams, denoted as `B4'. We also used both metrics for the ImgSp2Txt model in MMC2, which received image and speech inputs together. TTS was evaluated based on L2-norm$^2$ (denoted as `L2') error between ground-truth and predicted speech (lower being better), while IG was assessed with the inception score (IS) \cite{inception_score} to measure how realistic the IG output was (higher being better).
	
	\vspace{-0.25cm}
	\section{Experiment Results}
	\vspace{-0.10cm}
	\subsection{Large amount of paired data}
	\label{ssec:large}
	\vspace{-0.2cm}
	
	First, we evaluated the performance of the components when a large amount of paired data existed, as compared with the existing published results, on a well-known dataset. In this case, all components were trained independently using supervised training. However, in the previous publication of MMC1 \cite{effendi2019speech}, we showed that our ASR, TTS, and IC components could provide comparable performance to methods shown in other existing published papers. Since we used the same architectures and parameters for those components, we won't repeat the evaluation here. The main difference is that the current research uses IG and ImgSp2Txt. For the IG task, we assessed using the Caltech-UCSD Birds-200-2011 (CUB) dataset \cite{cub_dataset}, while
	for the ImgSp2Txt task, we evaluated the performance of our models on the Flickr8k set \cite{flickr8k_corpus}. Our IG model, which had a 5.67 inception score, performed better than AttnGAN \cite{attngan}, which had a score of only 4.36. Furthermore, our ImgSp2Txt results also exceeded those of Sun et al. \cite{sun_f8k}, who used a lattice rescoring algorithm. These results confirmed that in a fully supervised scenario, our models work well or even better than those from previously published papers.
	
	\vspace{-0.25cm}
	\subsection{Limited amount of paired data}
	\label{ssec:limited}
	\vspace{-0.2cm}
	
	As described in Sec.~\ref{ssec:overview} and Sec.~\ref{sec:setup}, given a limited amount of paired data, we first trained all model components independently in supervised training with paired units. In the label propagation method \cite{labelprop}, we used the initially trained models to generate the missing information in unpaired data (i.e., given speech data only, ASR produced the text transcription and created new pair data). After that, we retrained all models again independently using the new pair data with supervised learning. In contrast, in the proposed MMC1 and MMC2 frameworks, we used the initially trained models within a chain framework and enabled them to support each other, given unpaired data.

	Table~\ref{tab:result_audiovisual} shows the performance of the proposed MMC1 and MMC2 frameworks in comparison with the label propagation method in the Flickr8k dataset based on the training set-up listed in Table~\ref{tab:data}. `Label Propagation I' has the same amount of limited data with the proposed model, while `Label Propagation II' has more paired data for supervised learning. As this method could not be trained with mismatch data, we mark with ``-" in the table. The results reveal that the label propagation approach required more paired data to provide an improvement. Unfortunately, the increase is still minimal. On the other hand, the proposed MMC1 and MMC2 frameworks could give significantly better performance for all components than the label propagation method could.

	
	Among the proposed MMC1 and MMC2 frameworks, in the low-resource scenario, MMC2, with its more compact architecture using a single-loop mechanism, could mostly outperform MMC1 in all components except TTS. With a larger dataset, the CER result of MMC2 saturated in an on-par performance with MMC1. However, the BLEU score of MMC2 was much higher than that in MMC1. This was because of the joint use of image and speech as a multisource in the ImgSp2Txt model. The inception scores for the IG task in MMC2 were enhanced and were better than the scores for the IG task in MMC1. Interestingly, in both cases, ASR and TTS performance could still be improved and maintained using the image-only dataset, and IC and IG could even be improved using the speech-only dataset.

	\vspace{-0.2cm}
	\section{Related Works}
	\vspace{-0.2cm}
	
	Approaches that use learning from source-to-target and vice-versa as well as feedback links remain challenging. He et al. \cite{he2016dual} and Cheng et al. \cite{cheng_robust} proposed a mechanism called dual learning in neural machine translation. In image processing, several methods have also been proposed for unsupervised joint distribution matching without any paired data, such as DiscoGAN \cite{discogan}, CycleGAN \cite{CycleGAN2017}, and DualGAN \cite{dualgan}. The framework provides learning to translate an image from a source domain to a target domain without paired examples based on a cycle-consistent adversarial network. Implementation on voice conversion applications has also been investigated \cite{vc_cyclegan}. However, most of these works use only the same domain between the source and the target.
	
	The speech chain framework \cite{tjandra_schain4,tjandra_schain1,tjandra_schain2,tjandra_schain3} takes advantage of the duality of speech and text modalities by constructing a chain mechanism between ASR and TTS. For exploiting the duality between images and text, Huang et al. (2018) proposed turbo learning for joint training between image captioning and image generation \cite{huang_turbo}. Recently, a multimodal machine chain \cite{effendi2019speech} was proposed to accommodate multimodalities and a loop feedback mechanism. However, this work was only tested on a synthesized single-speaker dataset. In addition, this work is also unable to produce unseen images because the framework relied on IR model which only retrieve existing images.
	
	Specific to multimodal speech recognition tasks with deep learning, there are extensive studies \cite{petridis_av,chung2017lip,afouras2018deep,masr_2} that have attempted to combine both audio and visual information for better ASR. However, these studies mainly focused on lip visual information. Another work has addressed multimodality from a different perspective: using visual information as contextual information for a caption of something being spoken \cite{sun_f8k}. Unfortunately, this work was not implemented in an end-to-end manner. In this study, we constructed a multimodal chain with a single-loop and a dual-loop architecture that can be trained with semi-supervised learning.
	
	\vspace{-0.30cm}
	\section{Conclusion}
	\vspace{-0.2cm}
	
	We developed several new improvements in a multimodal machine chain. First, we improved the multimodal chain, MMC1, to handle unseen data better by incorporating an adversarial-based image generation model and speaker embedding for managing multispeaker variation. We also proposed an alternative multimodal chain, MMC2, using a single-loop architecture with a dual decoder and investigated the possibility of using audiovisual information when available. Significant improvement for all components was shown compared to the label generation method. The results reveal that both the MMC1 and MMC2 frameworks enable speech processing components to improve their performance when using an image-only dataset and image processing components to enhance their performance when using a speech-only dataset. For future work, we may further investigate the various approaches of component combination, not only for ImgSp2Txt but also for Txt2ImgSp.
	
	\vspace{-0.25cm}
	\section{Acknowledgements}
	\vspace{-0.2cm}
	Part of this work is supported by JSPS KAKENHI Grant Numbers JP17H06101 and JP17K00237 as well as NII CRIS Contract Research 2019 and the Google AI Focused Research Awards Program.
	
	\bibliographystyle{IEEEtran}
	\bibliography{mthesis}

\begin{thebibliography}{10}
\providecommand{\url}[1]{#1}
\csname url@samestyle\endcsname
\providecommand{\newblock}{\relax}
\providecommand{\bibinfo}[2]{#2}
\providecommand{\BIBentrySTDinterwordspacing}{\spaceskip=0pt\relax}
\providecommand{\BIBentryALTinterwordstretchfactor}{4}
\providecommand{\BIBentryALTinterwordspacing}{\spaceskip=\fontdimen2\font plus
\BIBentryALTinterwordstretchfactor\fontdimen3\font minus
  \fontdimen4\font\relax}
\providecommand{\BIBforeignlanguage}[2]{{%
\expandafter\ifx\csname l@#1\endcsname\relax
\typeout{** WARNING: IEEEtran.bst: No hyphenation pattern has been}%
\typeout{** loaded for the language `#1'. Using the pattern for}%
\typeout{** the default language instead.}%
\else
\language=\csname l@#1\endcsname
\fi
#2}}
\providecommand{\BIBdecl}{\relax}
\BIBdecl

\bibitem{denes1993speech}
P.~Denes and E.~Pinson, \emph{The Speech Chain}, ser. Anchor books.\hskip 1em
  plus 0.5em minus 0.4em\relax Worth Publishers, 1993.

\bibitem{tjandra_schain4}
A.~Tjandra, S.~Sakti, and S.~Nakamura, ``Machine speech chain,'' \emph{IEEE/ACM
  Transactions on Audio, Speech, and Language Processing}, vol.~28, pp.
  976--989, 2020.

\bibitem{tjandra_schain1}
------, ``Listening while speaking: Speech chain by deep learning,'' in
  \emph{Proc. of the IEEE ASRU}, Dec 2017, pp. 301--308.

\bibitem{tjandra_schain2}
------, ``Machine speech chain with one-shot speaker adaptation,'' in
  \emph{Proc. of INTERSPEECH}, 2018, pp. 887--891.

\bibitem{tjandra_schain3}
------, ``End-to-end feedback loss in speech chain framework via
  straight-through estimator,'' in \emph{Proc. of IEEE ICASSP}, 2019, pp.
  6281--6285.

\bibitem{effendi2019speech}
J.~Effendi, A.~Tjandra, S.~Sakti, and S.~Nakamura, ``Listening while speaking
  and visualizing: Improving {ASR} through multimodal chain,'' in \emph{Proc.
  of the 2019 IEEE ASRU}, 2019.

\bibitem{chan2016listen}
W.~Chan, N.~Jaitly, Q.~Le, and O.~Vinyals, ``Listen, attend and spell: A neural
  network for large vocabulary conversational speech recognition,'' in
  \emph{Proc. of IEEE ICASSP}, 2016, pp. 4960--4964.

\bibitem{wang_taco}
Y.~Wang, R.~Skerry-Ryan, D.~Stanton, Y.~Wu, R.~Weiss, N.~Jaitly, Z.~Yang,
  Y.~Xiao, Z.~Chen, S.~Bengio, Q.~Le, Y.~Agiomyrgiannakis, R.~Clark, and
  R.~Saurous, ``Tacotron: Towards end-to-end speech synthesis,'' in \emph{Proc.
  of the INTERSPEECH}, 2017, pp. 4006--4010.

\bibitem{xu2015show}
K.~Xu, J.~Ba, R.~Kiros, K.~Cho, A.~Courville, R.~Salakhudinov, R.~Zemel, and
  Y.~Bengio, ``Show, attend and tell: Neural image caption generation with
  visual attention,'' in \emph{Proc. of ICML}, 2015, pp. 2048--2057.

\bibitem{attngan}
T.~Xu, P.~Zhang, Q.~Huang, H.~Zhang, Z.~Gan, X.~Huang, and X.~He, ``Attn{GAN}:
  Fine-grained text to image generation with attentional generative adversarial
  networks,'' in \emph{Proc. of CVPR}, 2018.

\bibitem{he2016deep}
K.~He, X.~Zhang, S.~Ren, and J.~Sun, ``Deep residual learning for image
  recognition,'' in \emph{Proc. of IEEE CVPR}, 2016, pp. 770--778.

\bibitem{deepspeaker}
C.~Li, X.~Ma, B.~Jiang, X.~Li, X.~Zhang, X.~Liu, Y.~Cao, A.~Kannan, and Z.~Zhu,
  ``Deep speaker: an end-to-end neural speaker embedding system,'' \emph{CoRR},
  vol. abs/1705.02304, 2017.

\bibitem{flickr8k_corpus}
C.~Rashtchian, P.~Young, M.~Hodosh, and J.~Hockenmaier, ``Collecting image
  annotations using amazon's mechanical turk,'' in \emph{Proc. of the NAACL HLT
  2010 Workshop on Creating Speech and Language Data with Amazon's Mechanical
  Turk}, 2010, pp. 139--147.

\bibitem{flickr8k_audio}
D.~Harwath and J.~Glass, ``Deep multimodal semantic embeddings for speech and
  images,'' in \emph{Proc. of IEEE ASRU}, 2015, pp. 237--244.

\bibitem{sun_f8k}
F.~{Sun}, D.~{Harwath}, and J.~{Glass}, ``Look, listen, and decode: Multimodal
  speech recognition with images,'' in \emph{Proc. of IEEE SLT}, Dec 2016, pp.
  573--578.

\bibitem{labelprop}
X.~Zhu and Z.~Ghahramani, ``Learning from labeled and unlabeled data with label
  propagation,'' in \emph{Tech. Rep.}, 2002.

\bibitem{papineni2002bleu}
K.~Papineni, S.~Roukos, T.~Ward, and W.-J. Zhu, ``{BLEU}: a method for
  automatic evaluation of machine translation,'' in \emph{Proc. of the 40th
  annual meeting on association for computational linguistics}, 2002, pp.
  311--318.

\bibitem{inception_score}
T.~Salimans, I.~Goodfellow, W.~Zaremba, V.~Cheung, A.~Radford, and X.~Chen,
  ``Improved techniques for training gans,'' in \emph{Proc. of the advances in
  neural information processing systems (NIPS)}, 2016, pp. 2234--2242.

\bibitem{cub_dataset}
C.~Wah, S.~Branson, P.~Welinder, P.~Perona, and S.~Belongie, ``The
  {C}altech-{USCD} birds-200-2011 dataset,'' \emph{Computation and Neural
  Systems Technical Report}, 2011.

\bibitem{he2016dual}
D.~He, Y.~Xia, T.~Qin, L.~Wang, N.~Yu, T.-Y. Liu, and W.-Y. Ma, ``Dual learning
  for machine translation,'' in \emph{Proc. of NIPS}, 2016, pp. 820--828.

\bibitem{cheng_robust}
Y.~Cheng, Z.~Tu, F.~Meng, J.~Zhai, and Y.~Liu, ``Towards robust neural machine
  translation,'' in \emph{Proc. of ACL}, Melbourne, Australia, Jul. 2018, pp.
  1756--1766.

\bibitem{discogan}
T.~Kim, M.~Cha, H.~Kim, J.~K. Lee, and J.~Kim, ``Learning to discover
  cross-domain relations with generative adversarial networks,'' in \emph{Proc.
  of ICML}, 2017, pp. 1857--1865.

\bibitem{CycleGAN2017}
J.-Y. Zhu, T.~Park, P.~Isola, and A.~A. Efros, ``Unpaired image-to-image
  translation using cycle-consistent adversarial networks,'' in \emph{Proc. of
  IEEE ICCV}, 2017.

\bibitem{dualgan}
Z.~Yi, H.~Zhang, P.~Tan, and M.~Gong, ``Dual{GAN}: Unsupervised dual learning
  for image-to-image translation,'' in \emph{Proc. of ICCV}, 2017, pp.
  2868--2876.

\bibitem{vc_cyclegan}
K.~Tanaka, T.~Kaneko, N.~Hojo, and H.~Kameoka, ``Synthetic-to-natural speech
  waveform conversion using cycle-consistent adversarial networks,'' in
  \emph{Proc. of IEEE SLT}, 2018, pp. 632--639.

\bibitem{huang_turbo}
Q.~Huang, P.~Zhang, D.~O. Wu, and L.~Zhang, ``Turbo learning for captionbot and
  drawingbot,'' \emph{CoRR}, vol. abs/1805.08170, 2018.

\bibitem{petridis_av}
S.~Petridis, Y.~Wang, Z.~Li, and M.~Pantic, ``End-to-end audiovisual fusion
  with {LSTM}s,'' in \emph{Proc. of AVSP}, 2017, pp. 36--40.

\bibitem{chung2017lip}
J.~S. Chung, A.~Senior, O.~Vinyals, and A.~Zisserman, ``Lip reading sentences
  in the wild,'' in \emph{Proc. of IEEE CVPR}, 2017, pp. 3444--3453.

\bibitem{afouras2018deep}
T.~Afouras, J.~S. Chung, A.~Senior, O.~Vinyals, and A.~Zisserman, ``Deep
  audio-visual speech recognition,'' \emph{Proc. of IEEE transactions on
  pattern analysis and machine intelligence}, 2018.

\bibitem{masr_2}
H.~Cetingul, E.~Erzin, Y.~Yemez, and A.~Tekalp, ``Multimodal speaker/speech
  recognition using lip motion, lip texture and audio,'' \emph{Signal
  Processing}, vol.~86, no.~12, pp. 3549 -- 3558, 2006.

\end{thebibliography}
	
\end{document}